\documentclass[11pt,a4paper]{article}
\usepackage[
hyperref]{naaclhlt2019}
\usepackage{times}
\usepackage{latexsym}
\usepackage{microtype}
\usepackage{amsmath}
\usepackage{adjustbox}
\usepackage{tikz}
\usepackage{booktabs}
\usepackage[T1]{fontenc}
\usepackage[safe]{tipa}
\usepackage{inconsolata}
\usepackage{enumerate}
\usetikzlibrary{bayesnet}

\newcommand{\dech}{\mathbf{h}^{\textit{(dec)}}}
\newcommand{\ench}{\mathbf{h}^{\textit{(enc)}}}
\newcommand{\tagh}{\mathbf{h}^{\textit{(tag)}}}
\newcommand{\al}{\mathbf{a}}
\newcommand{\calA}{\mathcal{A}}

\usepackage{todonotes}
\makeatletter
\newcommand*\iftodonotes{\if@todonotes@disabled\expandafter\@secondoftwo\else\expandafter\@firstoftwo\fi}  %
\makeatother

\DeclareMathOperator*{\argmax}{argmax}

\usepackage{cleveref}
\crefname{section}{\S}{\S\S}
\Crefname{section}{\S}{\S\S}
\crefname{table}{Table}{}
\crefname{figure}{Figure}{}
\crefname{algorithm}{Algorithm}{}
\crefname{equation}{eq.}{}
\crefname{appendix}{App.}{}
\crefformat{section}{\S#2#1#3}  
\usepackage{url}

\newcommand{\word}[1]{\textit{#1}}
\newcommand{\defn}[1]{\textbf{#1}}
\newcommand{\vw}{\boldsymbol{w}}
\newcommand{\uw}{\mathbf{u}}

\newcommand{\vl}{{\boldsymbol \ell}}
\newcommand{\vm}{\boldsymbol{m}}
\newcommand{\Yc}{\mathcal{Y}}
\newcommand{\calK}{{\cal K}}
\aclfinalcopy %
\def\aclpaperid{805} %

\makeatletter
\newcommand{\printfnsymbol}[1]{%
  \textsuperscript{\@fnsymbol{#1}}%
}
\makeatother

\title{A Simple Joint Model for Improved Contextual Neural Lemmatization}

\author{Chaitanya Malaviya$^{\textrm{\normalfont*,\textipa{B}}}$ \and Shijie Wu$^{\textrm{\normalfont *,\normalfont \textschwa}}$ \and Ryan Cotterell$^{\textrm{\normalfont \textschwa,\textipa{H}}}$ \\
${}^{\textrm{\textipa{B}}}$Allen Institute for Artificial Intelligence\\
${}^{\textrm{\textschwa}}$Department of Computer Science, Johns Hopkins University \\
${}^{\textrm{\textipa{H}}}$Department of Computer Science and Technology, University of Cambridge  \\
{\tt \href{mailto:chaitanyam@allenai.org}{chaitanyam@allenai.org}}\quad{\tt \href{mailto:shijie.wu@jhu.edu}{shijie.wu@jhu.edu}}\quad{\tt \href{mailto:rdc42@cam.ac.uk}{rdc42@cam.ac.uk}}
}

\begin{document}
\maketitle
\begin{abstract}
English verbs have multiple forms. For instance, \word{talk} may also
appear as \word{talks}, \word{talked} or \word{talking}, depending on
the context. The NLP task of lemmatization seeks to map these diverse
forms back to a canonical one, known as the lemma. We present a simple
joint neural model for lemmatization and morphological tagging that
achieves state-of-the-art results on 20 languages from the Universal
Dependencies corpora. Our paper describes the model in addition to
training and decoding procedures. Error analysis indicates that joint
morphological tagging and lemmatization is especially helpful in
low-resource lemmatization and languages that display a larger degree
of morphological complexity. Code and pre-trained models are available at \url{https://sigmorphon.github.io/sharedtasks/2019/task2/}.
\end{abstract}

\section{Introduction}\label{sec:introduction}
\makeatletter
\def\blfootnote{\xdef\@thefnmark{}\@footnotetext}
\makeatother
\blfootnote{* Equal contribution. Listing order is random.}

Lemmatization is a core NLP task that involves a string-to-string
transduction from an inflected word form to its citation form, known
as the lemma. More concretely, consider the English sentence:
\word{The bulls are running in Pamplona}. A lemmatizer will seek to
map each word to a form you may find in a dictionary---for instance,
mapping \word{running} to \word{run}.  This linguistic normalization is important in several downstream NLP applications, especially for highly inflected languages. Lemmatization has previously been shown to improve recall for information retrieval \cite{kanis2010comparison, monz2001shallow}, to aid machine translation \cite{fraser2012modeling, chahuneau2013translating} and is a core part of modern parsing systems \cite{bjorkelund2010high,zeman2018conll}.

However, the task is quite nuanced as the proper choice of the lemma is context dependent. For instance, in the sentence \word{A running of the bulls took place in
  Pamplona}, the word \word{running} is its own lemma, since, here,
\word{running} is a noun rather than an inflected verb. Several counter-examples exist to this trend, as discussed in depth in \newcite{haspelmath2013understanding}. Thus, a good
lemmatizer must make use of some representation of each word's
sentential context. The research question in this work is, then, how
do we design a lemmatization model that best extracts the morpho-syntax from the
sentential context?\looseness=-1

\begin{figure}
  \centering
  \includegraphics[width=\columnwidth]{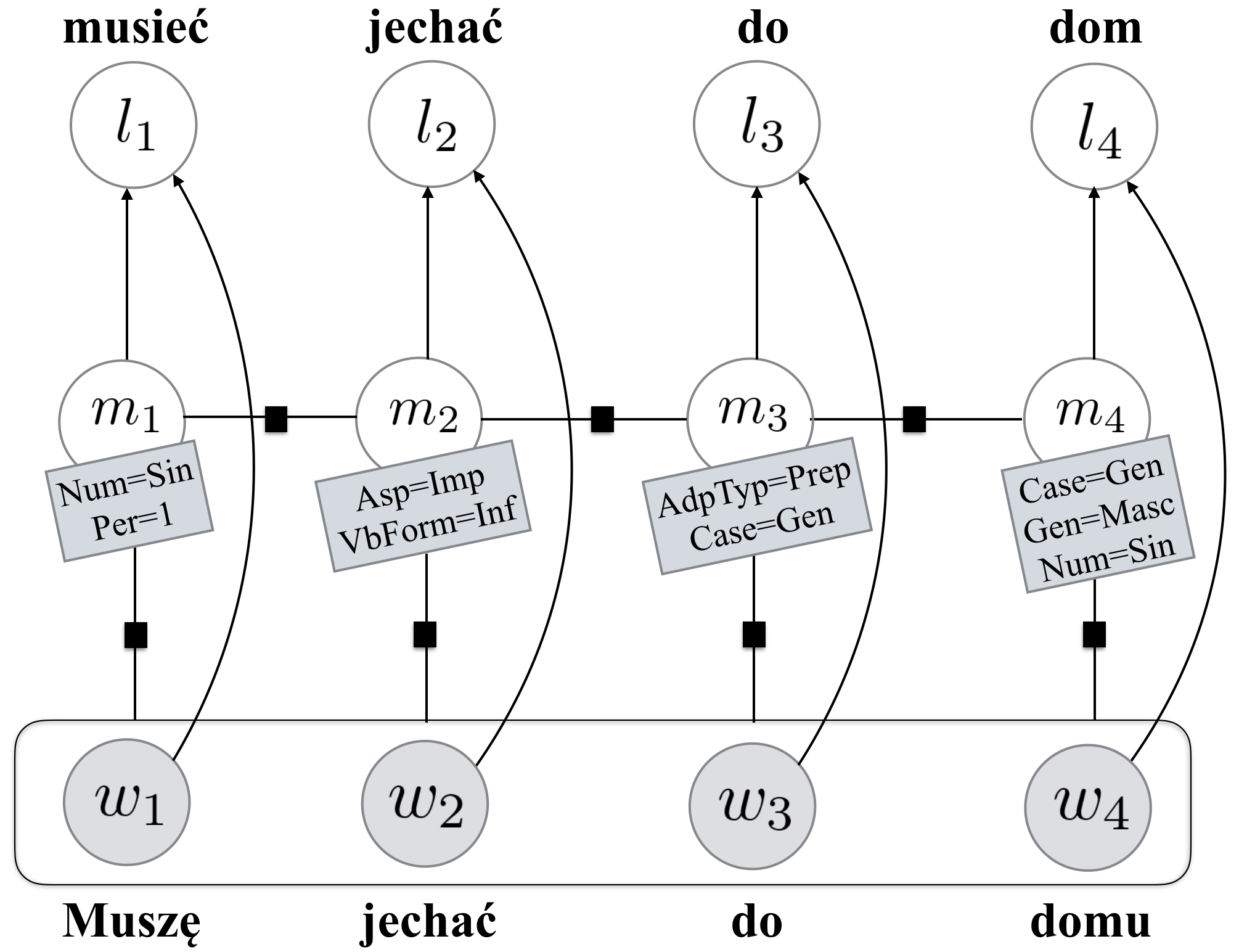}
  \caption{Our structured neural model shown as a hybrid
    (directed-undirected) graphical model
    \cite{koller2009probabilistic}. Notionally, the $w_i$ denote inflected word
    forms, the $m_i$ denote morphological tags and the $\ell_i$ denote
    lemmata.}
  \label{fig:model}
\end{figure}

\begin{figure*}
  \includegraphics[width=2\columnwidth]{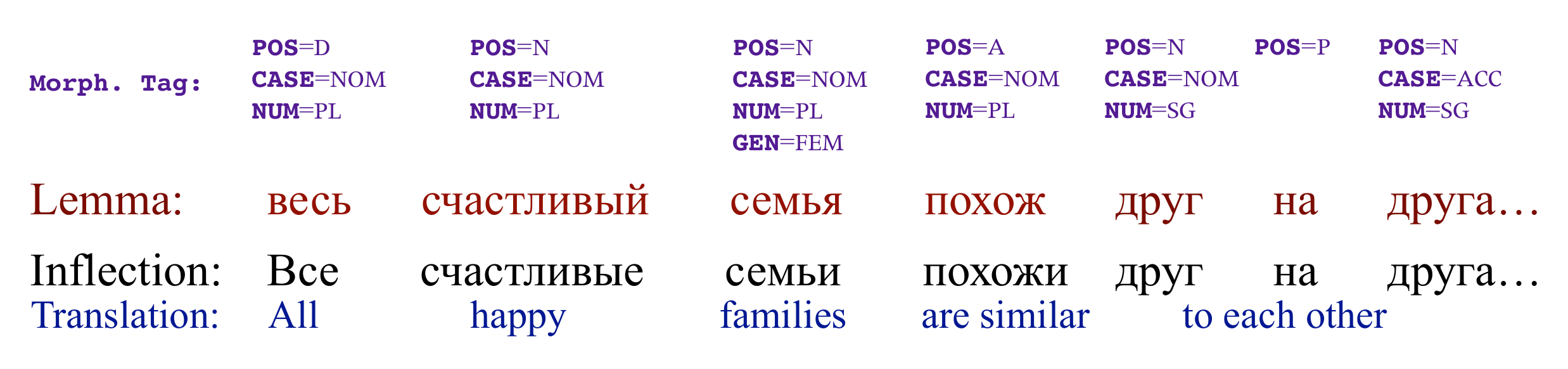}
  \caption{Example of a morphologically tagged (in purple) and lemmatized (in red) sentence in Russian using the annotation scheme provided in the UD dataset. The translation is given below (in blue).}
  \label{fig:sentence}
\end{figure*}

Recent work \cite{N18-1126} has presented a system that
directly summarizes the sentential context using a recurrent neural
network to decide how to lemmatize. As \newcite{N18-1126}'s
system currently achieves state-of-the-art results, it \emph{must} implicitly
learn a contextual representation that encodes the necessary
morpho-syntax, as such knowledge is requisite for the task. We contend, however, that rather than expecting the network
to \emph{implicitly} learn some notion of morpho-syntax, it is better
to \emph{explicitly} train a joint model to morphologically
disambiguate and lemmatize. Indeed, to this end, we introduce a joint model for the introduction of morphology into a neural lemmatizer.
A key feature of our model is its \emph{simplicity}: Our contribution is to show how to stitch existing models together into a joint model, explaining how to train and decode the model.
However, despite the model's simplicity,
it still achieves a significant improvement over the state of the art on our target task: lemmatization.\looseness=-1

Experimentally, our contributions are threefold. First, we show that
our joint model achieves state-of-the-art results, outperforming (on average) all competing approaches on a
20-language subset of the Universal Dependencies (UD) corpora
\cite{11234/1-1983}. Second, by providing the joint model with gold
morphological tags, we demonstrate that we are far from achieving the
upper bound on performance---improvements on morphological tagging
could lead to substantially better lemmatization. Finally, we provide
a detailed error analysis indicating \emph{when} and \emph{why}
morphological analysis helps lemmatization. We offer two
tangible recommendations: one is better off using a joint model (i)
for languages with fewer training data available and (ii)
languages that have richer morphology.

Our system and pre-trained models on all languages in the latest version of the UD corpora\footnote{We compare to previously published numbers on non-recent versions of UD, but the models we release are trained on the current version (2.3).}\footnote{Instead of UD schema for morphological attributes, we use the UniMorph schema \cite{sylak2016composition} instead. Note the mapping from UD schema to UniMorph schema is not one-to-one mapping \cite{mccarthy2018marrying}.} are released at \url{https://sigmorphon.github.io/sharedtasks/2019/task2/}.

\section{Background: Lemmatization}\label{sec:background}

Most languages \cite{wals} in the world exhibit a linguistic
phenomenon known as \defn{inflectional morphology}, which causes word
forms to mutate according to the syntactic category of the word. 
The syntactic context in which
the word form occurs determines which form is properly used.
One privileged form in the set of inflections is called the \defn{lemma}.
We regard the lemma as a lexicographic convention, often used to
better organize dictionaries. Thus, the choice of which inflected
form is the lemma is motivated by tradition and convenience, e.g., the
lemma is the infinitive for verbs in some Indo-European languages, rather
than by linguistic or cognitive concerns. Note that the \defn{stem}
differs from the lemma in that the stem may not be an actual
inflection.\footnote{The stem is also often ill-defined. What is, for
  instance, the stem of the word \word{running}, is it \word{run} or
  \word{runn}?} In the NLP literature, the syntactic category that
each inflected form encodes is called the \defn{morphological tag}. The morphological tag generalizes
traditional part-of-speech tags, enriching them with further
linguistic knowledge such as tense, mood, and grammatical
case. We call the individual key--attribute pairs \defn{morphological attributes}. An example of a sentence annotated with morphological tags and lemmata
in context is given in \cref{fig:sentence}. The task of mapping a
sentence to a sequence of morphological tags is known as
\defn{morphological tagging}.

\paragraph{Notation.}
Let $\vw = w_1,\ldots,w_N$ be a sequence of $N$ words. 
Each
individual word is denoted as $w_n$.  Likewise, let $\vm = m_1,\ldots,m_N$
and $\vl = \ell_1,\ldots,\ell_N$ be sequences of morphological tags
and lemmata, respectively. We will denote
the set of all tags seen in a treebank as $\Yc$. 
We remark that $m_n$ is $w_n$'s
morphological tag, e.g., $\left[\right.${\sc pos}$=${\sc n}, {\sc case}$=${\sc nom}, {\sc num}$=${\sc sg}$\left.\right]$ viewed as a single label, and $\ell_n$ is $w_n$'s lemma.
We will denote a language's discrete alphabet of characters as $\Sigma$. Thus, we have $w_n, \ell_n \in \Sigma^*$.\looseness=-1

\section{A Joint Neural Model}
The primary contribution of this paper is a joint model of
morphological tagging and lemmatization. 
The intuition behind the joint model is simple: high-accuracy lemmatization requires a
representation of the sentential context, in which the word occurs
(see the example given in \cref{sec:introduction}), and a
morphological tag provides a precise summary of the context
required to choose the correct lemma.
Armed with this intuition, we
define our joint model of lemmatization 
and morphological tagging as:
\begin{align}\label{eq:joint}
  p(\vl, \vm  \mid \vw) &= p(\vl \mid \vm, \vw)\,p(\vm \mid \vw) \\
                        &= \left(\prod_{n=1}^N  \underbrace{ p(\ell_n \mid m_n, w_n)}_\textit{meural transducer} \right)\, \underbrace{p(\vm \mid \vw)}_\textit{neural tagger} \nonumber
\end{align}
\Cref{fig:model} illustrates the structure of our model in the form of a graphical model. We will discuss the lemmatization factor and the morphological tagging factor
following two subsections, separately. We caution the reader that the discussion of these models will be brief: Neither of these particular components is novel with respect to the literature, so the formal details of the two models is best found in the original papers. The point of our paper is to describe a simple manner to combine these existing parts into a state-of-the-art lemmatizer.\looseness=-1

\subsection{Morphological Tagger: $p(\vm \mid \vw)$}
We employ a simple LSTM-based tagger to recover the morphological tags of a sentence \cite{E17-1048,D17-1078}. We also experimented with the neural conditional random field of \newcite{P18-1247}, but \newcite{E17-1048} gave slightly better tagging scores on average and is faster to train. Given a sequence of $N$ words $\vw = w_1 \cdots w_N$, we would like to obtain the morphological tags $\vm = m_1 \cdots m_N$ for each word, where $ m_n \in \Yc$. The model first obtains a word representation for each token using a character-level biLSTM \cite{graves2013speech} embedder, which is then input to a word-level biLSTM tagger that predicts tags for each word. Given a function cLSTM that returns the last hidden state of a character-based LSTM, first we obtain a word representation $\uw_n$ for word $w_n$ as:
\begin{equation}
\uw_n=[\textnormal{cLSTM}(c_1 \cdots c_M) ; \textnormal{cLSTM}(c_M \cdots c_1)]
\label{eq:charLSTM}
\end{equation}
where $w_n = c_1 \cdots c_M$ is the character sequence of the $n^{\text{th}}$ word.
This representation $\uw_n$ is then input into a word-level biLSTM tagger. 
The word-level biLSTM tagger predicts a tag from $\Yc$. 
A full description of the model is found in \newcite{E17-1048}.
We use the standard cross-entropy loss for training this model and decode greedily while predicting the tags during test-time.
Note that greedy decoding is optimal in this tagger as there is no interdependence between the tags $m_i$. 

\subsection{A Lemmatizer: $p(\ell_n \mid m_n, w_n)$}
Neural sequence-to-sequence models
\cite{DBLP:conf/nips/SutskeverVL14,bahdanau2014neural} have yielded
state-of-the-art performance on the task of generating morphological
variants---including the lemma---as evinced in several recent shared
tasks on the subject \cite{W16-2002,K17-2001,K18-3001}. Our lemmatization factor in
\cref{eq:joint} is based on such models. Specifically, we make use of
a hard-attention mechanism \cite{xu2015show,N16-1076}, rather than the original
soft-attention mechanism. Our choice of hard attention is motivated by
the performance of \citeposs{K18-3008} system at the CoNLL-SIGMORPHON
task. We use a nearly identical model, but opt for an exact dynamic-programming-based inference scheme \cite{D18-1473}.\footnote{Our formulation
  differs from the work of \newcite{D18-1473} in that we enforce monotonic
  hard alignments, rather than allow for non-monotonic alignments.}\looseness=-1

We briefly describe the joint model here.
Given an inflected word $w$ and a tag $m$, we
would like to obtain the lemma $\ell \in \Sigma^*$, dropping the subscripts for simplicity. 
We decompose the lemma into its constituent characters using the following notation $\ell = \kappa_1 \cdots \kappa_{L}$, where each $\kappa_j \in \Sigma$. 
A character-level biLSTM encoder embeds $w = c_1 \cdots c_M$ into a sequence of hidden states $\ench_1, \ldots, \ench_M$. 
The decoder LSTM produces
an embedding $\dech_j$, reading the concatenation of the embedding of at the previous time step $\dech_{j-1}$ and the tag embedding $\tagh$, which is produced by an order-invariant linear
function. 
In contrast to soft attention, hard attention models the alignment
distribution explicitly.\looseness=-1

We denote the set of all non-monotonic alignments from $w$ to $\ell$ as $\calA =\{1, \ldots, M\}^{L}$ and the notation $A_j = i$ refers to the
event that the $j^\text{th}$ character of $\ell$ is aligned to the $i^\text{th}$ character of $w$.
We factor the probabilistic lemmatizer as follows:
{
\allowdisplaybreaks
\begin{subequations}
\begin{align}
p(&\ell \mid m,w) = \sum_{\al \in \calA} p(\ell, \al \mid m,w)  \\
&= \sum_{\al \in \calA} \prod_{j=1}^{L} p(\kappa_j \mid A_j = a_j, \kappa_{< j}, m,w) \\
& \quad\quad \quad\quad\quad  \times p(a_j \mid a_{j-1}, \kappa_{< j}, m,w)  \nonumber \\
&= \sum_{\al \in \calA} \prod_{j=1}^{L} \,p(\kappa_j \mid \ench_{a_j},\dech_j) \\
& \quad\quad \quad\quad\quad \times p( a_j \mid a_{j-1},  \ench_{a_j},\dech_j) \nonumber
\end{align}
\end{subequations}
} \hspace{-.6cm} The summation is computed with dynamic programming---specifically, using the forward algorithm for hidden Markov models \cite{rabiner1989tutorial}.
The distribution $p(\kappa_j \mid \ench_{a_j},\dech_j)$ is a two-layer feed-forward network followed
by a softmax. 
The transition $p(a_j \mid a_{j-1},\ench_{a_{j-1}},\dech_j)$ is the multiplicative attention function with $\ench_{a_{j-1}}$ and $\dech_j$ as input. To
enforce monotonicity, $p(a_j \mid a_{j-1},\ench_{a_{j-1}},\dech_j) = 0$ if $a_j < a_{j-1}$. The exact details
of the lemmatizer are given by \newcite{wu-cotterell-2019-exact}.

\subsection{Decoding}
We consider two manners, by which we decode our model. The first
is a greedy decoding scheme. The second is a \defn{crunching}
\cite{N06-1045} scheme. We describe each in turn. 

\paragraph{Greedy Decoding.}
In the greedy scheme, we select the best
morphological tag sequence 
\begin{equation}
\vm^\star = \argmax_{\vm} \log p(\vm
\mid \vw)
\end{equation}
and then decode each lemmata 
\begin{equation}
\ell_n^\star =
\argmax_{\ell}p(\ell \mid m^\star_n, w_n) \label{eq:decode-lemmat}
\end{equation}
Note that we slightly abuse notation since the argmax here is \emph{approximate}: exact decoding of our neural lemmatizer is hard. This sort of scheme is also referred to as pipeline decoding.
\paragraph{Crunching.}
In the crunching scheme, we first extract
a $k$-best list of taggings from the morphological tagger.
For an input sentence $\vw$, call the $k$-best
tags for the $n^\text{th}$ word $\calK(w_n)$.
Crunching then says we should
decode in the following manner
\begin{align}
	\ell_n^\star =\argmax_{\ell} \log \!\!\!\! \sum_{m_n \in \calK(w_n)} \!\!\!\! p(\ell \mid m_n, w_n)\,p(m_n \mid \vw) 
\end{align}		
Crunching is a tractable heuristic that approximates
true joint decoding\footnote{True joint decoding
would sum over all possible morphological tags, rather than just the $k$-best. While this is tractable in our setting in the sense that there are, at, most, $1662$ morphological tags (in the case of Basque), it is significantly slower than using a smaller $k$. Indeed, the probability distribution that morphological taggers learn tend to be peaked to the point that considering improbable tags is not necessary.} and, as such, we expect
it to outperform the more na{\"i}ve greedy approach.

\subsection{Training with Jackknifing}
In our model, a simple application of maximum-likelihood estimation (MLE)
is unlikely to work well.
The reason is that our model is a discriminative directed
graphical model (as seen in \cref{fig:model}) and, thus, suffers
from \defn{exposure bias} \cite{ranzato2015sequence}. 
The intuition behind the poor
performance of MLE is simple: the output of the lemmatizer depends on the output
of the morphological tagger; as the lemmatizer has only ever seen correct morphological tags,
it has never learned to adjust for the errors that will be made at the time of decoding.
To compensate for this, we employ \defn{jackknifing} \cite{P17-2107},
 which is standard practice in many NLP pipelines, such as dependency parsing.

Jackknifing for training NLP pipelines is quite similar to the oft-employed cross-validation. We divide our training data into $\tau$ splits. Then, for each split $i \in \{1,\ldots, \tau\}$, we train the morphological tagger on \emph{all but} the $i^\text{th}$ split and then decode the trained tagger, using either greedy decoding or crunching, to get silver-standard tags for the held-out $i^{\text{th}}$ split. Finally, we take our
collection of silver-standard morphological taggings
and use those as input to the lemmatizer
in order to train it. 
This technique helps avoid exposure bias and improves the lemmatization performance, which we will demonstrate empirically in \cref{sec:exp}. Indeed, the model is quite ineffective without this training regime. Note that we employ jackknifing for both the greedy decoding scheme and the crunching decoding scheme. 

\section{Experimental Setup}\label{sec:exp}

\subsection{Dataset}\label{sec:dataset}
To enable a fair comparison with \newcite{N18-1126},
we use the Universal Dependencies Treebanks~\cite{11234/1-1983} for all our experiments. Following previous work, we use v2.0 of the treebanks for all languages, except Dutch, for which v2.1 was used due to inconsistencies in v2.0. The standard splits are used for all treebanks.

\subsection{Training Setup and Hyperparameters}\label{sec:training}
For the morphological tagger, we use the baseline implementation from \newcite{P18-1247}. This implementation uses an input layer and linear layer dimension of 128 and a 2-layer LSTM with a hidden layer dimension of 256. The Adam \citep{kingma2014adam} optimizer is used for training and a dropout rate \cite{srivastava2014dropout} of 0.3 is enforced during training. The tagger was trained for 10 epochs.\looseness=-1

For the lemmatizer, we use a 2-layer biLSTM encoder and a 1-layer LSTM decoder
with 400 hidden units. The dimensions of character and tag embedding are 200
and 40, respectively. We enforce a dropout rate of 0.4 in the embedding and encoder LSTM layers. The
lemmatizer is also trained with Adam and the learning rate is 0.001. We halve
the learning rate whenever the development log-likelihood increases and we perform early-stopping when the learning rate reaches $1\times10^{-5}$. We apply gradient clipping
with a maximum gradient norm of 5.\looseness=-1

\subsection{Baselines (and Related Work)}\label{sec:baselines}
Previous work on lemmatization has investigated both neural \cite{bergmanis_goldwater_NAACL2019} and non-neural \cite{chrupala2008towards,D15-1272,nicolai2016leveraging,K17-2001} methods. We compare our approach against competing
methods that report results on UD datasets.\looseness=-1
\paragraph{Lematus.} The current state of the art is held by \newcite{N18-1126}, who, as discussed in \cref{sec:introduction}, provide a direct context-to-lemma approach, avoiding the use of morphological tags. We remark that \newcite{N18-1126} assume a setting where lemmata are annotated at the token level, but morphological tags are not available; we contend, however, that such a setting is not entirely realistic as almost all corpora annotated with lemmata at the token level include morpho-syntactic annotation, including the vast majority of the UD corpora. Thus, we do not consider it a stretch to assume the annotation
of morphological tags to train our joint model.\footnote{After correspondence with Toms Bergmanis, we would like to clarify this point. While \newcite{N18-1126} explores the model in a token-annotated setting, as do we, the authors argue that such a model is better for a very low-resource scenario where
the entire sentence is not annotated for lemmata. We concede this point---our current model is not applicable in such a setting. However, we note that a semi-supervised morphological tagger could be trained in such a situation as well, which may benefit lemmatization.}  
\paragraph{UDPipe.} Our next baseline is
the UDPipe system of \newcite{K17-3009}. Their system performs lemmatization using an averaged perceptron tagger that predicts a (lemma rule, UPOS) pair. Here, a lemma rule generates a lemma by removing parts of the word prefix/suffix and prepending and appending a new prefix/suffix. A guesser first produces correct lemma rules and the tagger is used to disambiguate from them.

\paragraph{Lemming.} The strongest non-neural baseline we consider is the system of \newcite{D15-1272}, who, like us, develop a joint model of morphological tagging lemmatization. In contrast to us, however, their model is globally normalized \cite{laffertyCrf}. 
Due to their global normalization, they directly estimate the parameters of their model with MLE without worrying about exposure bias. However, in order to efficiently normalize the model, they heuristically limit the set of possible lemmata through the use of \defn{edit trees} \cite{chrupala2008towards}, which makes the computation of the partition function tractable.\looseness=-1

\begin{figure*}[t!]
\centering
  \includegraphics[width=2\columnwidth]{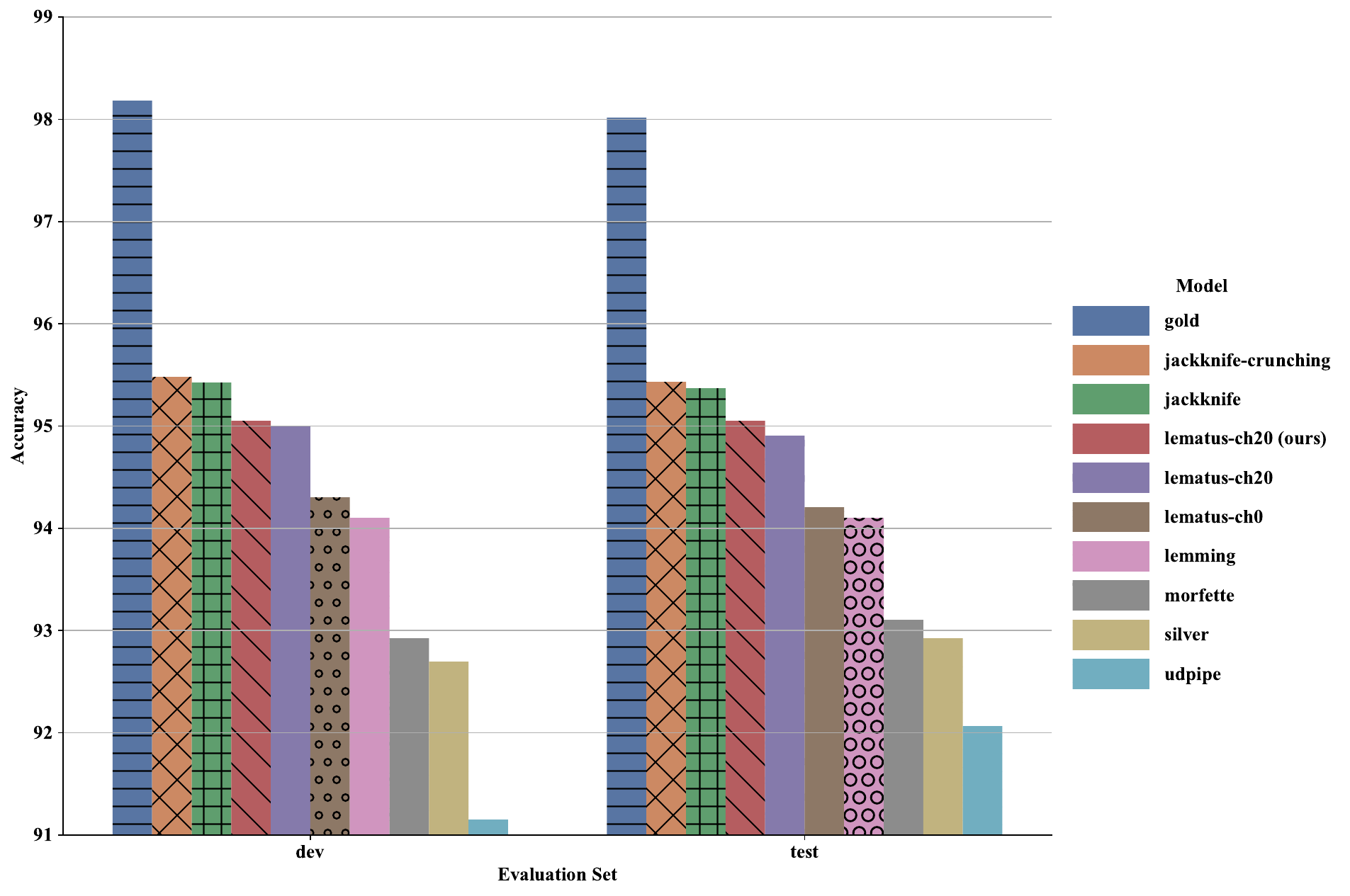}
  \caption{We present performance (in accuracy) averaged over the 20 languages from UD we consider. Our
    method (second from the left) significantly outperforms the strongest baseline (fourth from the left; \newcite{N18-1126}). The blue column is a skyline that gives our model gold tags during decoding, showing
  improved tagging should lead to better lemmatization. The remaining are baselines described in \cref{sec:baselines}.}
  \label{fig:results}
\end{figure*}

\paragraph{Morfette.}
Much like \newcite{D15-1272}, Morfette relies on the concept of edit trees. However, a simple perceptron
is used for classification with hand-crafted features. 
A full description of the model is given in \newcite{grzegorz2008learning}.

\section{Results and Discussion}\label{sec:results}
Experimentally, we aim to show three points.
i) Our joint model (\cref{eq:joint}) of morphological tagging
  and lemmatization achieves state-of-the-art accuracy; this
  builds on the findings of \newcite{N18-1126}, who show that
  context significantly helps neural lemmatization. Moreover, the upper bound for contextual lemmatizers that make use of
  morphological tags is much higher, indicating room for improved
  lemmatization with better morphological taggers.
ii) We discuss a number of error patterns that the model seems to make on the languages, where absolute accuracy is lowest: Latvian, Estonian and Arabic. We suggest possible paths forward to improve performance. 
iii) We offer an explanation for \emph{when} our joint model does better than the
  context-to-lemma baseline. We show in a correlational study that our joint approach with morphological tagging helps the most with low-resource and morphologically rich languages.

\subsection{Main Results}
The first experiment we run focuses on pure performance of the
model. Our goal is to determine whether joint morphological tagging
and lemmatization improves average performance in a state-of-the-art
neural model.

\paragraph{Evaluation Metrics.}
For measuring lemmatization performance, we measure the accuracy of guessing the lemmata correctly over an entire corpus. To demonstrate the effectiveness of our model in utilizing context and generalizing to unseen word forms, we follow \newcite{N18-1126} and also report accuracies on tokens that are i) \defn{ambiguous}, i.e., more than one lemmata exist for the same inflected form, ii) \defn{unseen}, i.e., where the inflected form has not been seen in the training set, and iii) \defn{seen unambiguous}, i.e., where the inflected form has only one lemma and is seen in the training set.\looseness=-1

\paragraph{Results.}
The results showing comparisons with all other methods are summarized in \cref{fig:results}. Additional results are presented in \cref{sec:additional-results}. Each
bar represents the average accuracy across 20 languages.
Our method achieves an average accuracy of $95.42$ and the strongest baseline,
\newcite{N18-1126}, achieves an average accuracy of $95.05$.
The difference in performance ($0.37$) is statistically significant with $p < 0.01$
under a paired permutation test.
We outperform the strongest baseline in 11 out of 20 languages and underperform in only 3 languages with $p<0.05$.
The difference between our method and all other baselines is statistical
significant with $p < 0.001$ in all cases. We highlight two additional features
of the data. First, decoding using gold morphological tags gives an accuracy of
$98.04$ for a difference in performance of $+2.62$. We take the large difference
between the upper bound and the current performance of our model to indicate
that improved morphological tagging is likely to significantly help
lemmatization. Second, it is noteworthy that training with gold tags, but
decoding with predicted tags, yields performance that is significantly worse
than every baseline except for UDPipe. This speaks for the importance of
jackknifing in the training of joint morphological tagger-lemmatizers that are
directed and, therefore, suffer from exposure bias.

\begin{figure}[]
\includegraphics[width=\columnwidth]{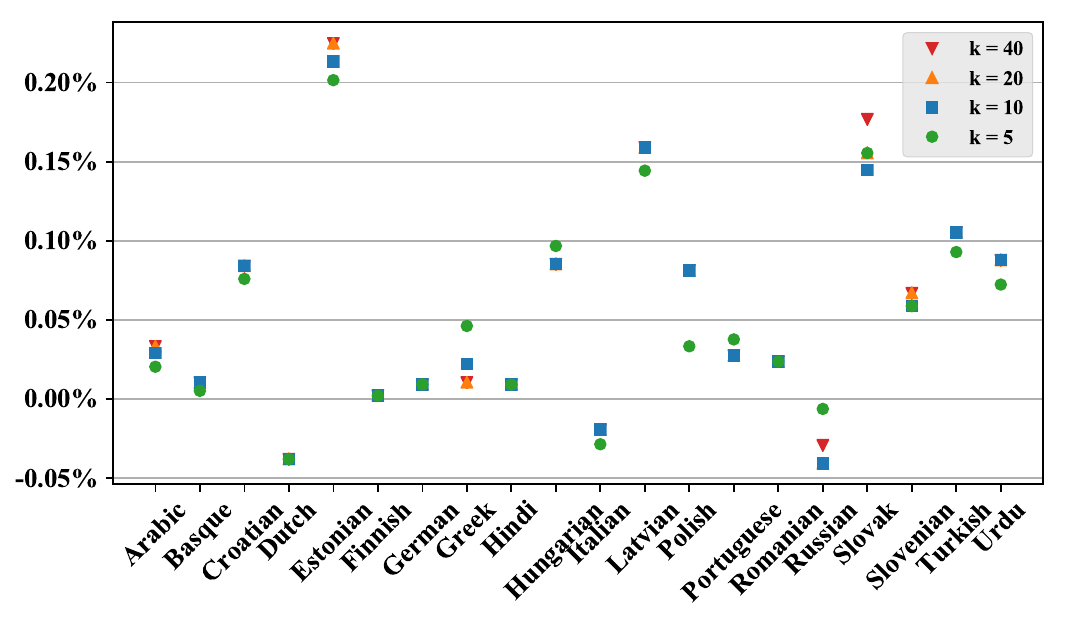}
\caption{Improvement on validation set with crunching over greedy decoding for different values of $k$.}
\label{fig:crunching}
\end{figure}

\begin{figure*}[]
\includegraphics[width=2\columnwidth]{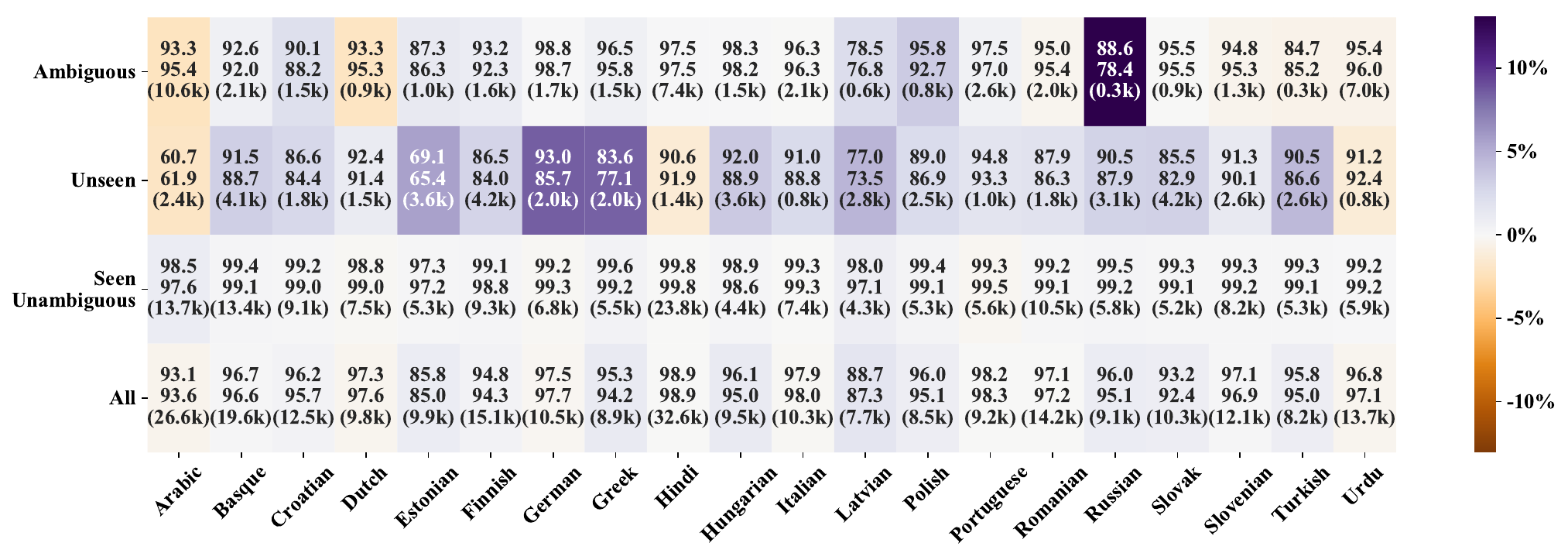}
\caption{Dev accuracy breakdown by type of inflected form on all languages comparing our system with greedy decoding against our run of Lematus-ch20, colored by relative improvement in percentage. In each entry, the bottom score is from Lematus-ch20 and the top one is from our system, and the number in the parenthesis is the number of tokens for the corresponding setting.}
\label{fig:error-analysis}
\end{figure*}

\paragraph{The Effect of Crunching.}
In \cref{fig:crunching}, we observed crunching further improves performance of the
greedy decoding scheme. In 8 out of 20 languages, the improvement is statistical
significant with $p < 0.05$. We select the best $k$ for each language based on the
development set.
In \cref{fig:error-analysis}, we provide a language-wise breakdown of the
performance of our model and the model of \newcite{N18-1126}. Our strongest
improvements are seen in Latvian, Greek and Hungarian. When measuring
performance solely over unseen inflected forms, we achieve even stronger gains
over the baseline method in most languages. This demonstrates the generalization
power of our model beyond word forms seen in the training set. In addition, our
accuracies on ambiguous tokens are also seen to be higher than the baseline on
average, with strong improvements on highly inflected languages such as Latvian
and Russian. Finally, on seen unambiguous tokens, we note improvements that
are similar across all languages.\looseness=-1

\subsection{Error Patterns}

We attempt to identify systematic error patterns of our model in an effort to motivate future work. For this analysis, we compare predictions of our model and the gold lemmata on three languages with the weakest absolute performance: Estonian, Latvian and Arabic. First, we note the differences in the average lengths of gold lemmata in the tokens we guess incorrectly and all the tokens in the corpus. The lemmata we guess incorrectly are on average 1.04 characters longer than the average length of all the lemmata in the corpus. We found that the length of the incorrect lemmata does not correlate strongly with their frequency. Next, we identify the most common set of edit operations in each language that would transform the incorrect hypothesis to the gold lemma. This set of edit operations was found to follow a power-law distribution.\looseness=-1

\begin{table}[tbh]
  \begin{adjustbox}{width=\columnwidth}
    \begin{tabular}{llllll} \toprule
      \textbf{lang} & \textbf{\# tokens} & \textbf{\# tags} & \textbf{ours} & \textbf{Lematus} & \textbf{$\Delta$} \\ \midrule
      arabic & 202000 & 349 & 93.1 & 93.55 & -0.48 \\
      basque & 59700 & 884 & 96.74 & 96.55 & 0.2 \\
      croatian & 146000 & 1105 & 96.16 & 95.7 & 0.48 \\
      dutch & 163000 & 62 & 97.26 & 97.65 & -0.4 \\
      estonian & 17000 & 482 & 85.83 & 84.99 & 0.99 \\
      finnish & 135000 & 1669 & 94.79 & 94.31 & 0.51 \\
      german & 227000 & 683 & 97.46 & 97.72 & -0.26 \\
      greek & 36500 & 346 & 95.29 & 94.22 & 1.13 \\
      hindi & 261000 & 939 & 98.88 & 98.92 & -0.05 \\
      hungarian & 16700 & 580 & 96.13 & 94.99 & 1.2 \\
      italian & 236000 & 278 & 97.93 & 98.04 & -0.11 \\
      latvian & 28000 & 640 & 88.67 & 87.31 & 1.56 \\
      polish & 52000 & 991 & 95.99 & 95.12 & 0.91 \\
      portuguese & 176000 & 375 & 98.2 & 98.26 & -0.06 \\
      romanian & 157000 & 451 & 97.11 & 97.19 & -0.08 \\
      russian & 58400 & 715 & 96.0 & 95.07 & 0.98 \\
      slovak & 64800 & 1186 & 93.25 & 92.43 & 0.89 \\
      slovenian & 96500 & 1101 & 97.07 & 96.9 & 0.17 \\
      turkish & 31200 & 972 & 95.81 & 95.01 & 0.85 \\
      urdu & 101000 & 1001 & 96.76 & 97.12 & -0.37 \\ \bottomrule
  \end{tabular}
  \end{adjustbox}
  \caption{Here we present the number of tokens in each of the UD treebanks and the number of morphological tags. We take the number of tags as a proxy for the morphological complexity of the language. We also present numbers on the validation set with greedy decoding and from Lematus. Correlations between the
  first two columns are shown in \cref{tab:correlations}.}
  \label{tab:diffs}
  
\end{table}

For the case of Latvian, we find that the operation \{\textit{replace}: s $\rightarrow$ a\} is the most common error made by our model. This operation corresponds to a possible issue in the Latvian treebank, where adjectives were marked with gendered lemmas. This issue has now been resolved in the latest version of the treebank.
For Estonian, the operation  \{\textit{insert}: m, \textit{insert}: a\} is the most common error. The suffix \textit{-ma} in Estonian is used to indicate the infinitive form of verbs. Gold lemmata for verbs in Estonian are marked in their infinitive forms whereas our system predicts the stems of these verbs instead. These inflected forms are usually ambiguous and we believe that the model doesn't generalize well to different form-lemma pairs, partly due to fewer training data available for Estonian. This is an example of an error pattern that could be corrected using improved morphological information about the tokens. Finally, in Arabic, we find that the most common error pattern corresponds to a single ambiguous word form, \textit{'an} , which can be lemmatized as \textit{'anna} (like ``that'' in English) or \textit{'an} (like ``to'' in English) depending on the usage of the word in context. The word \textit{'anna} must be followed by a nominal sentence while \textit{'an} is followed by a verb. Hence, models that can incorporate rich contextual information would be able to avoid such errors.

\subsection{Why our model performs better?}
Simply presenting improved results does not entirely satiate our
curiosity: we would also like to understand \emph{why} our model
performs better. Specifically, we have assumed an additional level of
supervision---namely, the annotation of morphological tags.
We provide the differences between our method and our retraining
of the Lematus system presented in \cref{tab:diffs}. In addition to the performance
of the systems, we also list the number of tokens in each treebank
and the number of distinct morphological tags per language.
We perform a correlational study, which is shown in
\cref{tab:correlations}.

\paragraph{Morphological Complexity and Performance.}
We see that there is a moderate positive
correlation ($\rho=0.209$) between the number of morphological tags in a language
and the improvement our model obtains. As we take
the number of tags as a proxy for the morphological complexity in the
language, we view this as an indication that attempting to directly
extract the relevant morpho-syntactic information from the corpus
is not as effective when there is more to learn. In such languages,
we recommend exploiting the additional annotation to achieve better results.\looseness=-1

\begin{table}
  \begin{tabular}{lll} \toprule
    & Pearson's $R$ & Spearman's $\rho$ \\ \midrule
    \textbf{\# tags vs. $\Delta$} & 0.206 & 0.209 \\
\textbf{\# tokens vs. $\Delta$} &  -0.808 & -0.845 \\ \bottomrule
  \end{tabular}
  \caption{The table shows the correlations between the differences in dev performance between our model with greedy decoding and Lematus and two aspects of the data: number of tokens and number of tags.} \label{tab:correlations}
\end{table}

\begin{figure}
  \centering
  \includegraphics[width=\columnwidth]{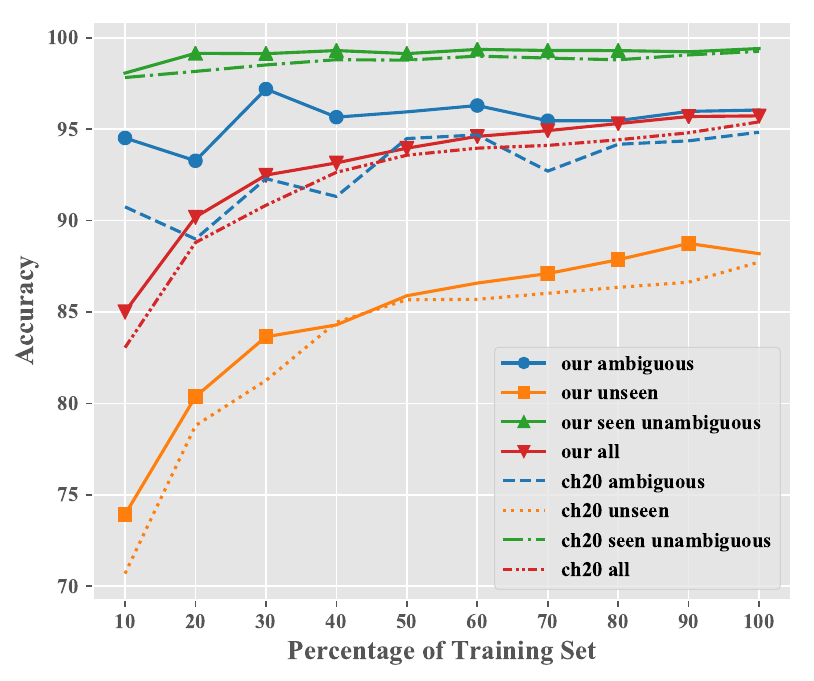}
  \caption{Learning curve showing the accuracy on the validation set of the Polish treebank as the percentage of training set is increased. Markers indicate statistically significant better system with paired permutation test ($p < 0.05$). Our model is decoded greedily.}
  \label{fig:learning}
\end{figure}

\paragraph{Amount of Data and Performance.}
The second correlation we find is a stronger negative correlation
($\rho=-0.845$) between the number of tokens available for training
in the treebank and the gains in performance of our model over
the baseline. This is further demonstrated by the learning curve plot in \cref{fig:learning}, where we plot the validation accuracy on the Polish treebank for different sizes of the training set. The gap between the performance of our model and Lematus-ch20 is larger when fewer training data are available, especially for ambiguous tokens. This indicates that the incorporation of morphological tags into a model helps more in the low-resource
setting. Indeed, this conclusion makes sense---neural networks
are good at extracting features from text when there is a sufficiently large amount of data. However, in the low-resource case, we would expect direct supervision on the sort of features we desire to extract to work better. Thus, our second
recommendation is to model tags jointly with lemmata when fewer training tokens are available. As we noted earlier, it is almost always the case that token-level annotation
of lemmata comes with token-level annotation of morphological tags. In low-resource scenarios, a data augmentation approach such as the one proposed by \citet{bergmanis_goldwater_NAACL2019} can be helpful and serve complementary to our approach.

\section{Conclusion}
We have presented a simple joint model for morphological tagging and lemmatization
and discussed techniques for training and decoding. 
Empirically, we have shown that our model achieves state-of-the-art results, hinting
that explicitly modeling morphological tags is a more effective manner for modeling context. In addition to strong numbers, we tried to explain \emph{when} and \emph{why} our model does better. Specifically, we show a significant correlation between our scores and the number of tokens and tags present in a treebank. We take this to indicate that our method improves performance more for low-resource languages as well as morphologically rich languages.

\section*{Acknowledgments}

We thank Toms Bergmanis for his detailed feedback on the accepted version of the manuscript. Additionally, we would like to thank the three anonymous reviewers for their valuable suggestions. The last author would like to acknowledge support from a Facebook Fellowship. 

\bibliography{naaclhlt2019}
\bibliographystyle{acl_natbib}

\appendix
\newpage~\newpage
\section{Additional Results}\label{sec:additional-results}

We present the exact numbers on all languages to allow future papers to compare to our results in \cref{tab:dev} and \cref{tab:test}. We also
present morphological tagging results in \cref{tab:morph}.

\begin{table}[tbh]
\begin{adjustbox}{width=1\columnwidth}
\begin{tabular}{l lllll}
\toprule
 & Gold & Crunching & Jackknifing & Ch-20 & Silver \\
\midrule
Arabic & 97.40 & 93.13 & 93.10 & 93.55 & 89.32 \\
Basque & 98.70 & 96.75 & 96.74 & 96.55 & 94.88 \\
Croatian & 98.61 & 96.24 & 96.16 & 95.70 & 94.89 \\
Dutch & 98.82 & 97.26 & 97.26 & 97.65 & 96.31 \\
Estonian & 91.92 & 86.02 & 85.83 & 84.99 & 71.65 \\
Finnish & 97.01 & 94.79 & 94.79 & 94.31 & 90.89 \\
German & 97.74 & 97.47 & 97.46 & 97.72 & 96.27 \\
Greek & 97.40 & 95.33 & 95.29 & 94.22 & 94.33 \\
Hindi & 99.29 & 98.89 & 98.88 & 98.92 & 98.63 \\
Hungarian & 98.54 & 96.22 & 96.13 & 94.99 & 94.15 \\
Italian & 99.37 & 97.93 & 97.93 & 98.04 & 96.95 \\
Latvian & 97.24 & 88.81 & 88.67 & 87.31 & 85.58 \\
Polish & 98.46 & 96.07 & 95.99 & 95.12 & 91.77 \\
Portuguese & 99.63 & 98.24 & 98.20 & 98.26 & 97.72 \\
Romanian & 99.34 & 97.13 & 97.11 & 97.19 & 95.99 \\
Russian & 98.46 & 96.00 & 96.00 & 95.07 & 93.84 \\
Slovak & 97.14 & 93.41 & 93.25 & 92.43 & 88.49 \\
Slovenian & 99.46 & 97.13 & 97.07 & 96.90 & 95.64 \\
Turkish & 98.71 & 95.91 & 95.81 & 95.01 & 91.25 \\
Urdu & 97.48 & 96.84 & 96.76 & 97.12 & 96.62 \\ \midrule
\textbf{AVERAGE} & 98.04 & 95.48 & 95.42 & 95.05 & 92.76 \\
\bottomrule
\end{tabular}
\end{adjustbox}
\caption{Development performance breakdown.}
\label{tab:dev}
\end{table}

\begin{table}[tbh]
\begin{adjustbox}{width=1\columnwidth}
\begin{tabular}{l lllll}
\toprule
 & Gold & Crunching & Jackknifing & Ch-20 & Silver \\
\midrule
Arabic & 97.95 & 93.99 & 93.92 & 94.16 & 91.37 \\
Basque & 98.54 & 96.63 & 96.67 & 96.49 & 94.57 \\
Croatian & 98.24 & 95.63 & 95.58 & 95.22 & 94.28 \\
Dutch & 98.43 & 97.25 & 97.25 & 97.21 & 96.50 \\
Estonian & 92.34 & 86.33 & 86.13 & 85.44 & 73.41 \\
Finnish & 97.02 & 94.34 & 94.29 & 93.94 & 90.57 \\
German & 97.39 & 97.14 & 97.07 & 97.63 & 95.88 \\
Greek & 97.83 & 96.53 & 96.46 & 95.32 & 95.05 \\
Hindi & 99.10 & 98.68 & 98.65 & 98.73 & 98.47 \\
Hungarian & 97.72 & 94.02 & 93.96 & 93.15 & 92.42 \\
Italian & 99.33 & 97.83 & 97.83 & 98.05 & 96.96 \\
Latvian & 96.69 & 89.79 & 89.76 & 88.87 & 86.49 \\
Polish & 98.45 & 95.74 & 95.78 & 94.90 & 91.94 \\
Portuguese & 99.60 & 97.97 & 97.86 & 98.14 & 97.58 \\
Romanian & 99.55 & 97.21 & 97.14 & 97.21 & 96.36 \\
Russian & 98.30 & 95.82 & 95.82 & 94.77 & 93.71 \\
Slovak & 97.40 & 93.46 & 93.31 & 92.29 & 88.63 \\
Slovenian & 99.25 & 96.74 & 96.66 & 96.69 & 95.42 \\
Turkish & 99.18 & 96.48 & 96.32 & 95.99 & 92.14 \\
Urdu & 97.87 & 96.94 & 96.91 & 96.77 & 96.73 \\ \midrule
\textbf{AVERAGE} & 98.01 & 95.43 & 95.37 & 95.05 & 92.92 \\
\bottomrule
\end{tabular}
\end{adjustbox}
\caption{Test performance breakdown.}
\label{tab:test}
\end{table}

\begin{table}[t!]
\begin{tabular}{ll}
\toprule
 & F1 Score \\
\midrule
Arabic & 85.62 \\
Basque & 83.68 \\
Croatian & 85.37  \\
Dutch & 90.92  \\
Estonian & 65.80 \\
Finnish & 87.94  \\
German & 79.45 \\
Greek & 87.63 \\
Hindi & 87.89 \\
Hungarian & 86.00 \\
Italian & 93.78 \\
Latvian & 80.96 \\
Polish & 80.29 \\
Portuguese & 93.65 \\
Romanian & 93.51 \\
Russian & 83.69 \\
Slovak & 64.53 \\
Slovenian & 88.81 \\
Turkish & 82.60 \\
Urdu & 72.86 \\ \midrule
\textbf{AVERAGE} & 83.75 \\
\bottomrule
\end{tabular}
\caption{Morphological Tagging Performance on development set.}
\label{tab:morph}
\end{table}

\end{document}